## Semantic, Cognitive, and Perceptual Computing: Advances toward Computing for Human Experience

Amit Sheth and Pramod Anantharam, Kno.e.sis, Wright State University
Cory Henson, Bosch Research and Technology Center

While the debate about whether AI, robots, or machines will replace humans is raging (among Gates, Hawking, Musk, Thiel, and others), there remains a long tradition of viewpoints that take a progressively more human-centric view of computing. A range of viewpoints from machine-centric to human-centric computing have been put forward by McCarthy (intelligent machines) [JM07], Weiser (ubiquitous computing) [MW91], Engelbart (augmenting human intellect) [DE62], Licklider (man-machine symbiosis) [JCRL60], and others as shown in Figure 1. In this paper, we focus on the recent progress taking place in the tradition of human-centric computing, such as Computing for Human Experience (CHE) (Sheth) [S10] and Experiential Computing (Jain) [RJ03]. CHE focuses on serving our needs, empowering us while keeping us in the loop, making us more productive with better and timelier decision-making, and improving and enriching our quality of life. Experiential Computing proposes to utilize the symbiotic relationship between computers and people and exploit their relative strengths, of symbolic/logic manipulation and complex pattern recognition respectively.

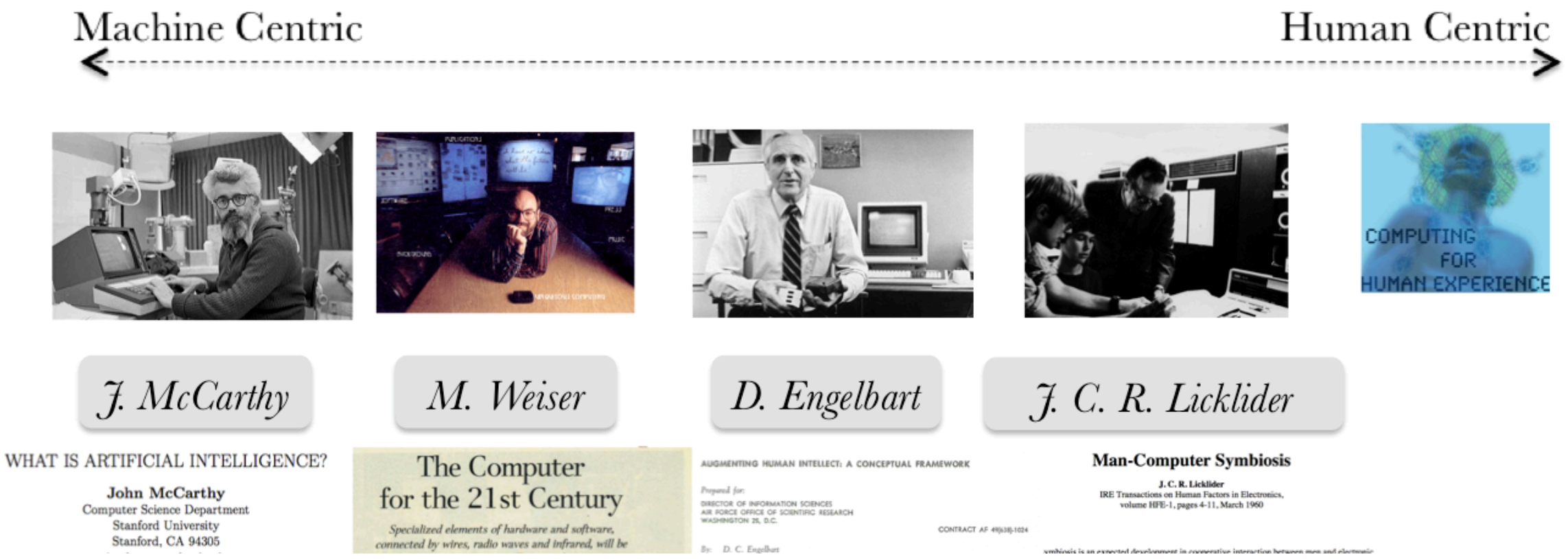

**Figure 1 A wide gamut of computing extending from machine centric to human centric**

CHE utilizes the World Wide Web (Web) to manage and share massive amounts of multimodal and multisensory observations; observations which capture the moments of people's lives. This includes various situations pertinent to people's needs and interests, along with some of their idiosyncrasies. Data of relevance to people's lives span the physical, cyber, and social spheres [SAH13]. The physical sphere encompasses reality, as measured by sensors/devices/Internet of Things; the cyber sphere encompasses all shared data and knowledge on the Web, and the social sphere encompasses all the human interactions and conversations. Observation data on the Web may represent events of interest to a population of people (e.g., climate), to a sub-population (e.g., traffic), or to an individual (something very personal, like an asthma attack). These observations contribute toward shaping the human experience[1], which is defined as the materialization of feelings, beliefs, facts and ideas that can be acted upon.

CHE emphasizes a *contextual* and *personalized* interpretation of data which is more readily consumable and *actionable* for people. Toward this goal, we discuss the

[1] based on: http://arachnoid.com/levels/ , also definition 1 of http://www.merriam-webster.com/dictionary/experience

computing paradigms of semantic computing, cognitive computing, and an emerging paradigm in this lineage, which we term *perceptual computing*. We believe that these technologies offer a continuum that reaches toward the goal of making the most of the vast, growing, and diverse data about things that matter to people's needs, interests, and experiences. This is achieved through actionable information, both when humans desire something (explicit action) and through ambient understanding (implicit action) of when something may be useful to people's activities and decision-making. Perceptual computing, in particular, is characterized by its use of *interpretation* and *exploration* to actively interact with the surrounding environment in order to collect data of relevance and usefulness for understanding the world around us.

This article consists of two parts. First we describe semantic computing, cognitive computing, and perceptual computing to draw distinctions while acknowledging their complementary capabilities to support CHE. We then provide a conceptual overview of the newest of these three paradigms—perceptual computing. For further insights, we describe a scenario of asthma management and explain the computational contributions of semantic computing, cognitive computing, and perceptual computing in synthesizing actionable information. This is done through computational support for the contextual and personalized processing of data into abstractions that move it closer to the level of the human comprehension and decision-making.

## 1. Challenge: Making the Web More Intelligent to Serve People Better

As we continue to make progress in developing technologies that disappear into the background, as envisaged by Mark Weiser [MW91], the next important focus of *human-centered computing* is to endow the Web, and computing in general, with sophisticated, human-like capabilities to reduce information overload. In the near future, computers will be able to process and analyze data, in a highly contextual and personalized manner, at a scale much larger than the human brain is able to handle. This technology will provide more intimate support to our every decision and action, ultimately shaping the human experience. The three capabilities we consider include: *semantics*, *cognition*, and *perception*. While dictionary definitions for cognition and perception often have significant overlap, we will make a distinction based on how cognitive computing has been defined thus far and on the complementary capabilities of *perceptual computing*. Over the next decade the development of these three computing paradigms—both individually and in cooperation—and their integration into the fabric of the Web will enable the emergence of a far more intelligent and human-centered Web.

## 2. Semantics, Perception, and Cognition

Semantics, perception, and cognition have been defined and utilized in a variety of ways. We would like to clarify our interpretation of these terms and also specify the connections between them in the context of human cognition and perception, and how observational data relates to semantics, perception, and cognition. Accordingly, we ignore their use in other contexts; for example, the use of semantics in the context of programming languages, or perception in the context of people interacting with computing peripherals, which is more closely associated with Human Computer Interaction. Although we provide a brief overview of human cognition and perception to provide a broader context, we cannot do justice to summarizing related work in such a broad topic area. Our focus, in this paper, is on *computing paradigms* inspired by cognition and perception.

**Semantics** is the meaning attributed to concepts and their relationships within the mind. This network of concepts and relations is used to represent knowledge about the world [AK58]. Such knowledge may then be utilized for interpreting our daily experiences through cognition and perception. Semantic concepts may represent (or unify, or subsume, or map to) various patterns of data, e.g., an observer may recognize a person by her face (visual signal) or by her voice (speech signal). Once recognized, however, both the visual and speech signals represent a single semantic concept of a person as recognized by the observer. Semantics hide the syntactic and representational differences in data and helps refer to data using a conceptual abstraction. Generally, this involves mapping observations from various physical stimuli, such as visual or speech signals, to concepts and relationships as humans would interpret and communicate them.

**Perception** is an act of interpreting data from the world around us. Perception involves pattern recognition and the classification of patterns from sensory inputs generated from physical stimuli, resulting in the formation of feelings, beliefs, facts, and ideas. Perception utilizes both sensing and actuation to actively explore the surrounding environment in order to collect data of relevance. This data may then be utilized by our cognitive facilities to more effectively understand the world around us. Perception involves *interpretation* and *exploration* with a strong reliance on background knowledge [G97].

Thus, perception is a cyclical process of interpretation and exploration of data utilizing associated knowledge of the domain. Perception constantly attempts to match incoming sensory inputs (and engage associated cognition) with top-down expectations or predictions (based on cognition) and closely integrates with human actions. In the context of asthma, *control level* is synthesized by various fine-grained observations (coughing, activity level, disturbed sleep) from patients, possibly involving iterative mechanism for seeking more information. Doctors and patients can easily comprehend the notion of control level, which is an abstraction resulting from perception.

**Cognition** is an act of understanding of the world around us by utilizing all the data from perception, with the context provided by existing knowledge. This understanding is achieved by utilizing domain/background knowledge and reasoning [DRSAAR11]. Understanding is highly contextual and personalized; for example, a doctor may interpret reduced activity differently for patients with varying asthma severity levels. A doctor understands reduced activity as a serious symptom for a patient with mild asthma while reduced activity is normal for a patient with severe asthma. Cognition enables contextual and personalized understanding of data from perception.

*Proposed/optional callout box*

*Semantics, perception, and cognition* interact seamlessly. Semantics makes an observation or data meaningful (i.e., provides a definition within the context of a system or knowledge of people), which in turn allows processing through relating or integrating with other observations and data. While the outcome of cognition results in understanding of our environment, the act of perception results in applying our understanding for exploring our environment. *Cognition enables perception to explore the most promising exploration path by providing a comprehensive understanding through the incorporation of background knowledge*

## 3. Computational Aspects of Semantics, Cognition, and Perception

For conceptual clarity and general understanding of what the three terms mean, we exemplify semantics, cognition, and perception using a real-world scenario of asthma management. Asthma is a *multifaceted*, highly *contextual*, and *personal* disease. Asthma is *multifaceted* since many aspects such as environmental triggers and patient sensitiveness to these triggers characterize it. Asthma is *contextual* since events of interest, such as the location of the person and triggers at a location, are crucial for timely alerts. Asthma is *personal* since asthma patients have varying responses to triggers and their actions vary based on the severity of their condition.

Asthma patients are characterized by two measures as used by clinical guidelines: severity level and control level. The severity level is diagnosed by a doctor and can take one of four states: *mild*, *mild persistent*, *moderate*, and *severe*. The control level indicates the extent of asthma exacerbations and can take one of three states: *well controlled*, *moderately controlled*, and *poorly controlled*. Patients do not exhibit change in their severity level often, but their control level may vary drastically depending on triggers, environmental conditions, medication, and symptomatic variations of the patient.

Let's consider an asthma patient, Anna, who is 10 years old and has been diagnosed with 'severe' asthma. She is taking her medication consistently and avoids exposure to triggers, resulting in a control level of 'well controlled'. She receives an invitation to play soccer in a few days. Now, Anna and her parents must maintain a balance between her wanting to play in soccer and the need to avoid exacerbating her asthma.

A solution to this dilemma is not straightforward, and cannot be found using only existing factual information found on the Web, in medical books or journals, or Electronic Medical Records (EMR). Knowledge found on the Web may contain common knowledge about asthma, but it may not be directly applicable to Anna. For example, while there may be websites [ATM15] describing general symptoms, triggers, and tips for better asthma management, Anna and her parents may not be able to rely on this information since her symptomatic variations for environmental triggers may be unique. While medical domain knowledge of asthma in the form of publications (e.g., PubMed) may contain symptomatic variations for various triggers, it is challenging to apply this knowledge to Anna's specific case, even though it may be described in her EMR. Furthermore, such an application of knowledge would not consider any environmental and physiological dynamics of Anna or her quality of life choices.

With the intention to provide more specifics, we will explore the semantic, cognitive, and perceptual computing aspects and then explain their role in providing a solution to the asthma control.

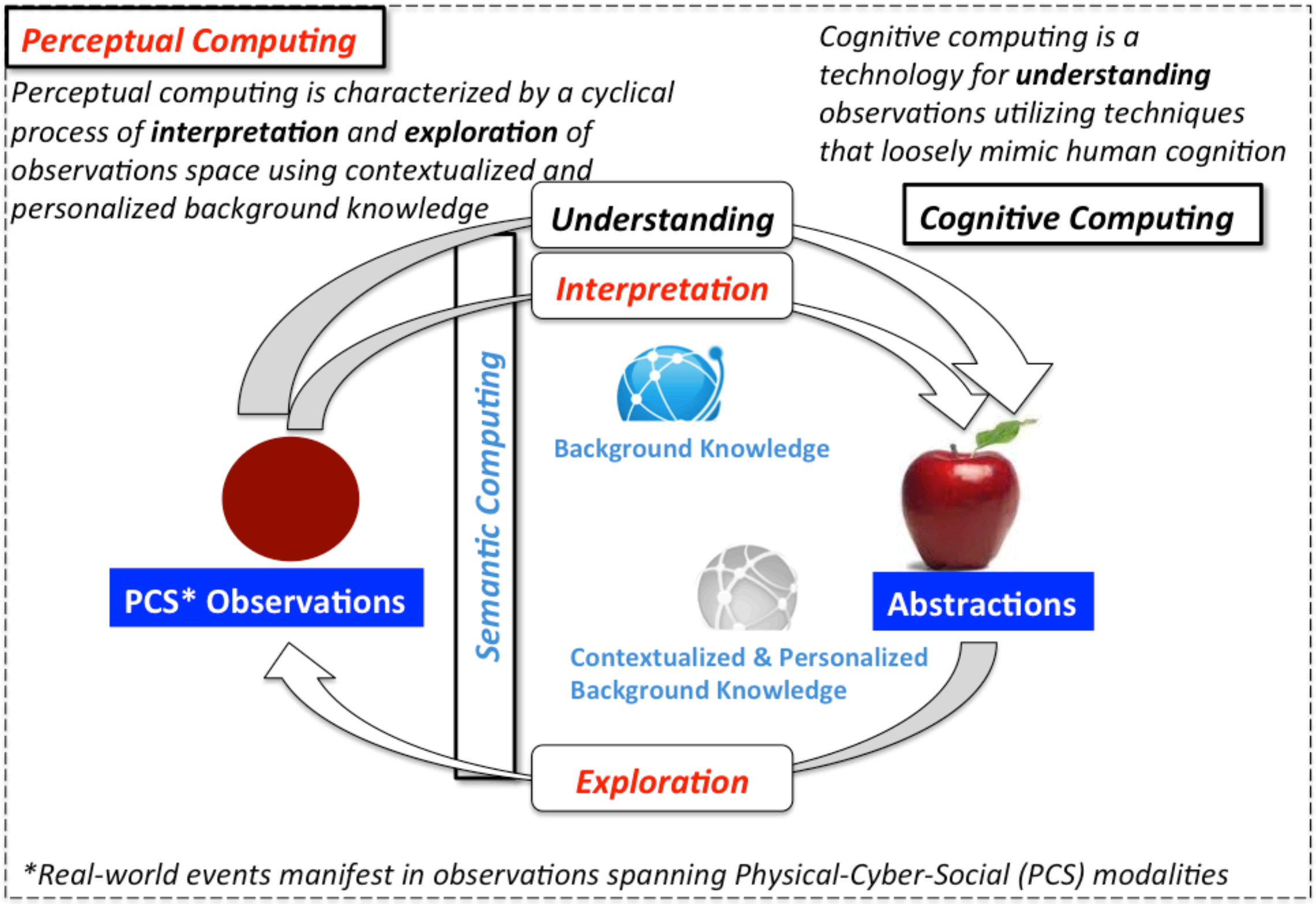

**Figure 2 Conceptual distinctions between perceptual, cognitive, and semantic computing along with a demonstration of the cyclical process of perceptual computing, which utilizes and refines background knowledge to include contextualization and personalization**

### 3.1 Semantic Computing (SC)

SC encompasses technology for representing concepts and their relations in an integrated semantic network that loosely mimics the inter-relation of concepts in the human mind. This conceptual knowledge, represented formally in an ontology, can be used to annotate data and infer new knowledge from interpreted data (e.g., to infer expectations of recognized concepts). Additionally, SC plays a crucial role in dealing with multisensory and multimodal observations, leading to the integration of observations from diverse sources (see "horizontal operators" in [SAH13]). SC has a rich history spanning 15 years [S11], and resulting in various annotation standards for a variety of data (e.g., social and sensor data [C12] are in use). The annotated data is used for interpretation by cognitive and perceptual computing. Figure 2 has SC as a vertical box through which interpretation and exploration are routed (further explained in Section 3.3). SC also provides languages for the formal representation of background knowledge.

The semantic network of general medical domain knowledge related to asthma and its symptoms define asthma control levels in terms of symptoms. This general knowledge may be integrated with knowledge of Anna's specific case found in her EMR. The weather, pollen, and air quality index information observed by sensors may be available through web services. These annotated observations spanning multiple modalities, general domain knowledge, and context-specific knowledge (Anna's asthma severity and control level) pose a great challenge for interpretation. The interpretation of observations needs background knowledge and, unfortunately, Anna's parents do not possess such asthma-related knowledge. Anna and her parents are left with no particular insights at this stage since manually interpreting all the observations is not a practical solution. In the next two subsections, we describe the interpretation of data using domain knowledge for deriving deeper insights.

### 3.2 Cognitive Computing (CC)

DARPA, when launching a project on cognitive computing in 2002 [DARPA02], defined cognitive computing as "reason[ing], [the] use [of] represented knowledge, learn[ing] from experience, accumulat[ing] knowledge, explain[ing] itself, accept[ing] direction, be[ing] aware of its own behavior and capabilities as well as respond[ing] in a robust manner to surprises." Cognitive hardware architectures and cognitive algorithms are two broad focus areas of current research in CC. Cognitive algorithms interpret data by learning and matching patterns in a way that loosely mimics the process of cognition in the human mind. There are various efforts in understanding human cognition in terms of computation [TCTN-11]. Cognitive systems learn from their experiences and then get better when performing repeated tasks. CC acts as prosthetics for human cognition by analyzing a massive amount of data and being able to answer questions humans may have when making certain decisions. One such example is IBM Watson, which won the game show Jeopardy! in early 2011. The IBM Watson approach (albeit, not the technology) is now extended to medicine to aid doctors in clinical decisions. CC interprets annotated observations obtained from SC, or raw observations from diverse sources, and presents the results to humans. Humans, in turn, can utilize the interpretation to perform an action, which forms additional input for the CC system. CC systems utilize machine learning and other AI techniques to achieve this, without being explicitly programmed. Figure 2 shows the role of CC in the interpretation and understanding of observations by utilizing background knowledge.

We will examine the role of CC in the asthma management scenario of Anna. Bewildered by the challenges in making the decision about whether she can play soccer, Anna's parents contact Anna's pediatrician, Dr. Jones, for help. Let's assume that Dr. Jones has access to a CC system such as IBM Watson for medicine [W15] and specifically for asthma management. Consequently, Dr. Jones is assisted by a CC system that can analyze massive amounts of medical literature, EMR, and clinical outcomes for asthma patients. Such a system would be instrumental in extending the cognitive abilities of Dr. Jones (minimizing the cognitive overload caused by the ever-increasing amounts of research literature). Dr. Jones discovers from the medical literature and EMRs that people with well-controlled asthma (i.e., patients who match Anna's control level) can indeed engage in physical activities if under the influence of appropriate preventive medication. Dr. Jones, however, is still unclear about the vulnerability of Anna's asthma control level due to weather and air quality index fluctuations. Dr. Jones lacks personalized and contextualized knowledge about Anna's day-to-day environment, rendering him ill-informed to make any recommendation to Anna.

### 3.3 Perceptual Computing (PC)[2]

Socrates taught that knowledge is attained through the careful and deliberate process of asking and answering questions. Through data mining, pattern recognition, and natural language processing, CC is rapidly progressing towards developing technology to support our ability to answer complex questions. PC will complete the loop by providing

[2] PC is referred to as an evolving paradigm of computing, and it is different from Perceptual Computing SDK by Intel which includes support for better human computer interaction

a technology to support our ability to ask contextually relevant and personalized questions [HST12]. PC complements SC and CC by providing machinery to ask the next question or derive a hypothesis based on observations, help identify what additional facts and observations can help evaluate or refine the hypothesis, in turn aiding decision-makers in gaining actionable insights. In other words, determining what data is most relevant in helping to disambiguate between the multiple possible causes (of Anna's asthma condition, for example). If the expectations derived by utilizing domain knowledge and observations from the real world do not match the real-world outcomes, PC updates the knowledge of the real world. Through focused attention, utilizing sensing and actuation technologies, this relevant data is sought in the environments spanning the Physical domain, consisting of sensor/IoT, the Cyber domain, including Web-based data/information/knowledge such as Wikipedia and Linked Open Data, and Social domain, including user-generated data.

PC envisions a more effective interpretation of data through a cyclical process of interpretation and exploration in a way that loosely mimics the process of perception in the human mind and body. Neisser defines perception as "an active, cyclical process of exploration and interpretation" [N67]. Machine perception is the process of converting sensor observations to abstractions through a cyclical process of interpretation and exploration utilizing background knowledge [HST12]. While CC efforts to date have investigated the interpretation of data, it has yet to adequately address the relationship between the interpretation of data and the exploration of (or interaction with) the environment. Additionally, PC involves the highly personalized and contextualized management (including additions and refinement) *and application of background knowledge* by engaging in the cyclical process of interpretation and exploration.

According to the theory of cognitive models [SM15], the bottom-brain organizes the received signals from our sense organs resulting in our perception of the real world. The top-brain deals with planning, goal setting, and even deals with dynamically changing goals and outcomes. Interpretation is analogous to bottom-brain operation of processing observations from our senses and exploration compares to the top-brain processing of making/adapting plans to solve problems [K13]. This type of interaction—often involving focused attention and physical actuation—enables the perceiver to collect and retain data of relevance (from the ocean of all possible data), and thus it facilitates a more efficient, personalized interpretation or abstraction.

Figure 2 demonstrates the cyclical process of PC involving interpretation and exploration. The interpretation of observations leads to abstractions (a concept in the background knowledge) and exploration leads to actuation to seek the most relevant next observation (to disambiguate between possible abstractions). SC implements horizontal operators for semantic integration of multimodal and multisensory observations [SAH13]. SC is also characterized by explicit modeling of the domain, though reasoning can be implicit. CC implements vertical operators [SAH13] for generating ranked hypothesis for a question explicitly asked by a person by utilizing massive amounts of unstructured data. A CC system facilitates cyclical interaction between people and the computing system for constant learning and improvement of the generated hypothesis (answers) to questions. The interaction is explicit i.e., initiated by people resulting in a symbiotic relationship between people and machines. PC implements horizontal operators for integration of heterogeneous and multimodal observations. PC also implements vertical operators [SAH13] for transforming massive amounts of multimodal and multisensory observations into abstractions intelligible to people. In addition to technologies used by SC and CC, PC focuses on machine perception [CH13] for exploration and interpretation of environment and observations, respectively. The cyclical process of exploration and interpretation is mostly implicit (when background knowledge is available) but may switch to explicit mode for incorporating inputs from people.

We adapt Figure 2, which presents conceptual distinctions between SC, CC, and PC to the asthma scenario, to further exemplify the role of each of the three computing paradigms. Figure 3 provides the observations and abstractions specific to the asthma management scenario. Background knowledge contains generic knowledge of asthma, such as the fact that asthma control level depends on the number of nights of disturbed sleep in a week, number of days of coughing in a week, and number of days of reduced activity in a week. This information can be obtained from a CC system [SS14] which analyzes asthma literature and scholarly articles and journals to answer a question such as: What constitutes the asthma control level? A CC system helps us in understanding the symptoms and their thresholds for asthma. There are some unanswered questions, such as: What does reduced activity mean in terms of number of steps taken per day? What does disturbed sleep mean in terms of the duration of Rapid Eye Movement (REM) sleep per night? Personalization of  asthma knowledge and learning normal levels of activity and sleep patterns has to be done before answering these two questions. PC enables personalization of generic background knowledge through the iterative cycle of interpretation and exploration to learn normalcy for a person. The abstractions indicating the asthma control level is much more intelligible to doctors for recommending corrective actions.

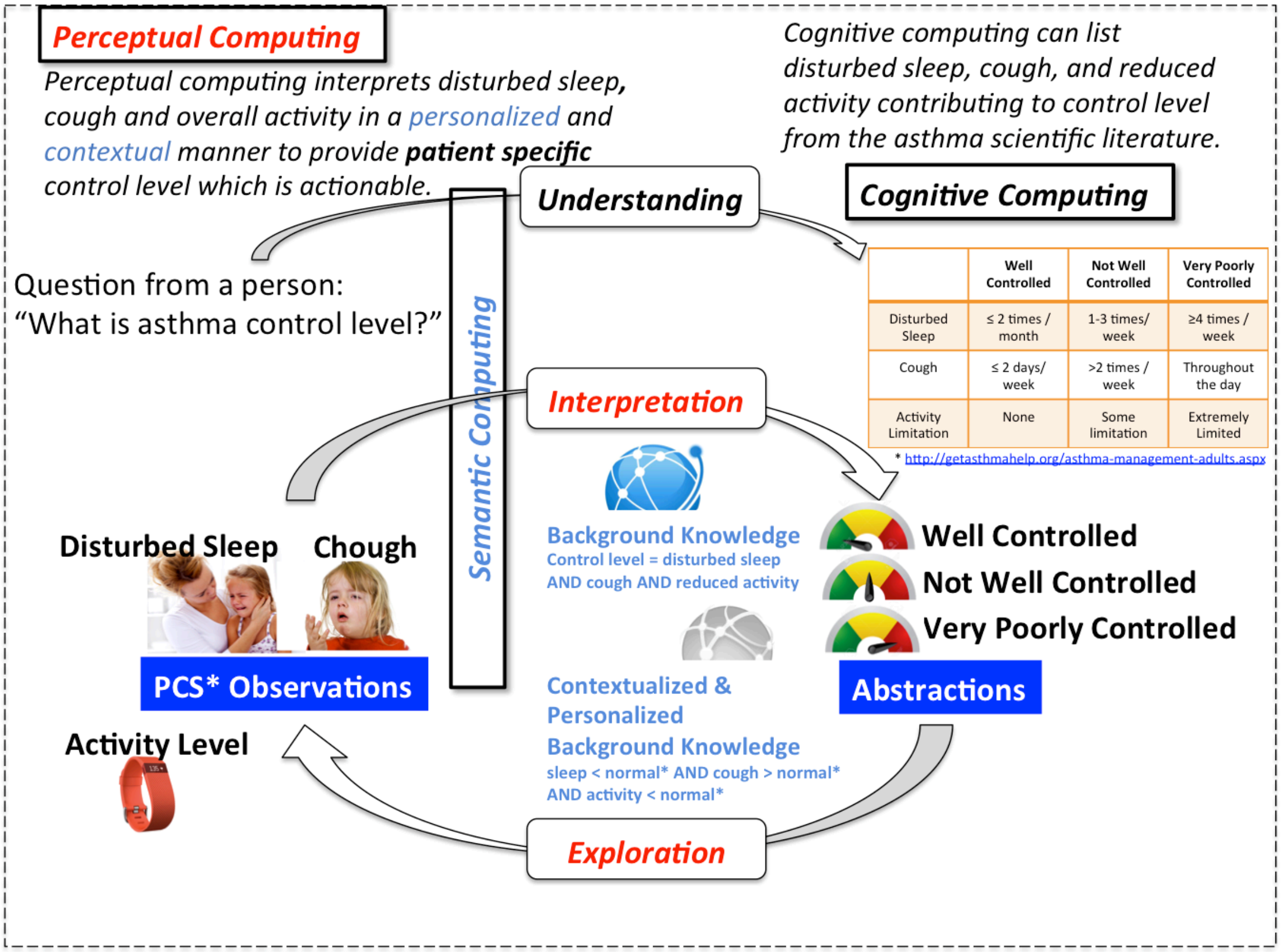

Figure 3 Asthma scenario to demonstrate the role of SC, CC, and PC for providing actionable information

For clarity, we provide an end-to-end example starting from the collected data and demonstrate the capabilities of SC, CC, and PC as shown in Figure 4. SC makes raw data more meaningful by annotating data with semantic concepts defined in an ontology. Consequently, SC adds meaning to data for enhanced consumption, reasoning, and sharing. The SSN ontology [C12, LL11] defines concepts and relationships for modeling sensors and their observations. In Figure 4, there are three observation types: sleep quality, number of steps, and the number of times the person coughed in a day. These raw data points do not carry much meaning, e.g., 1 hour 17 minutes, 672, and 20. This raw data can be enriched by linking them through annotation to concepts defined in an ontology; in this case, REM sleep, steps, and cough incidents respectively. Annotated data is amenable to knowledge-aware interpretation, which is a valuable feature, especially in knowledge-rich domains such as medicine. SC provides a language to represent such concepts and allows for the linking of raw data points to concepts in the ontology as shown in the annotation step of Figure 4. Nevertheless, the annotated data does not have sufficient direct value to the doctor i.e., Dr. Jones cannot use the annotated data for recommending any action to Anna.

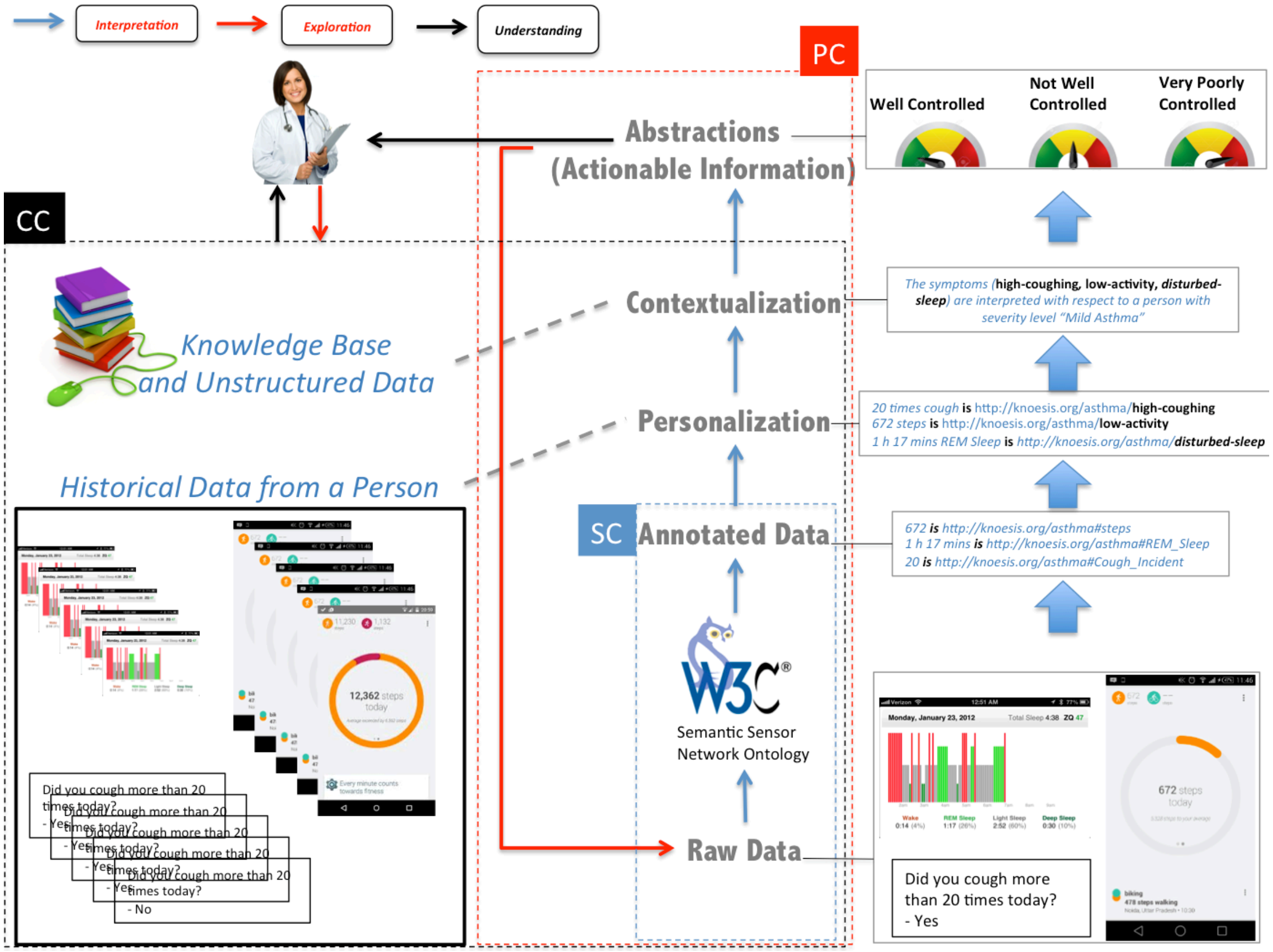

Figure 4 Operations performed over raw data by SC, CC, and PC for the asthma management scenario

A CC system with access to asthma articles and medical journals (unstructured data) and historical data from the patient can provide information on treatment regimes, medications, risks, and patient outcomes. A CC system can reveal valuable information that is otherwise hidden in massive amounts of medical literature to doctors, alleviating the problems they face in keeping up with the ever-expanding array of medical literature. Doctors need to apply this information to the specific context of a patient using their experience of past patients and their symptomatic variations. Doctors don't often have direct access to patient specific normalcy in terms of sleep, activity, and symptoms. Therefore, they need to learn this information from patients through a series of questions. In the asthma scenario, Dr. Jones may not get accurate information from Anna or her parents on sleep, activity, and coughing. Even if Dr. Jones manages to get some information from Anna or her parents, there is no supporting evidence for the obtained information. Dr. Jones has to deal with a lot of uncertainty, forcing him to base his recommendations on educated guesses, which is not ideal especially due to the contextual and personal variability of asthma symptoms.

A PC system significantly reduces the effort of a decision maker, such as Dr. Jones, who faces such challenges. A PC system is capable of consuming annotated data for personalization and contextualization as shown in Figure 4. Personalization involves the historical data of the patient for deriving information of interest in relation to the disease. Personalization by PC results in deriving patient-specific normalcy, a challenge for Dr. Jones while only utilizing a CC based system. Dr. Jones now has access to personalized data from Anna synthesized by utilizing normalcy information and the interpretation of current observations. Personalization for Anna would categorize the annotated data of sleep, activity, and coughing into disturbed-sleep, low-activity, and high coughing. Interpreting disturbed-sleep, low-activity, and high coughing depends on the asthma severity level of the person experiencing these symptoms. The asthma control level assigned to the person is conditioned on both symptoms and the asthma severity level. PC does this contextual interpretation to provide abstractions. These abstractions can be understood by decision makers, such as Dr. Jones, and used to provide a recommendation to Anna. The decisions and recommendations are now based on evidence provided by the PC system rather than educated guesswork.

With the rise of mobile computing and IoT technologies, it may be necessary that a PC system can be implemented as an ***intelligence at the edge*** technology [HTS12]***,*** as opposed to a logically-centralized system that processes the massive amounts of data on the Web. Since the computation can be carried out on a mobile device, the data remains on the local device allowing for better user control of data access, sharing, and privacy. In the scenario of Anna, a CC system processed all the medical knowledge, EMRs, and patient outcomes to provide information to Dr. Jones who then applied it to Anna's case (personalization). Dr. Jones faced challenges in interpreting weather data and the air quality index with respect to the vulnerability of Anna's asthma control level. With the PC system running on a local device (closer to Anna, possibly realized as a mobile application with inputs from multiple sensors, such as kHealth[3]), it can actively engage in the cyclical process of ***interpretation*** and ***exploration***. For example, Anna in the last month exhibited reduced activity during a soccer practice. This observation is interpreted by a PC system as an instance of asthma exacerbation. Further, a PC system actively seeks observations (asking questions by a PC system) of weather and the air quality to determine their effect on the asthma symptoms of Anna. A PC system will be able to take generic background knowledge (poor air quality *may cause* asthma exacerbations) for exploration and add contextual and personalized knowledge (poor air quality exposure of Anna *may cause* asthma exacerbations to Anna). Dr. Jones can be granted access to this information along with the information from the CC component. Anna is advised to refrain from the soccer match due to poor air quality on the day of the soccer game. This information will be valuable to Anna and her parents, possibly

[3] http://wiki.knoesis.org/index.php/Asthma

resulting in Anna avoiding situations that may lead to an exacerbation of her asthma condition.

## 4. Conclusions

This article outlines and exemplifies synergy between three important and complementary computing paradigms SC, CC, and PC. SC is perhaps best understood, with substantial technological support, and provides the ability to deal with the challenges of big data. CC has garnered substantial recent interest and provides the ability to utilize relevant knowledge and help improve the understanding of data for decision-making, and is seeing rapid technological progress. PC, a paradigm that is now being defined and understood, has the potential to bring tremendous value through its ability to provide personalized and contextual abstractions over massive amounts of multimodal data, originating from the physical, cyber, and social domains.

Using SC, CC, and PC synergistically, computers can not only provide answers to the complex questions posed to them but can also subsequently ask the right follow up questions and interact with the environment—either physical, cyber, or social—to collect the relevant data. As PC evolves, the personalization components will extend to include temporal and spatial context in addition to other factors that drive human decisions and actions, such as emotions and cultural/social preferences. This will enable more effective answers, better decisions, and more timely actions that are specifically tailored to each person. We envision the cyclical process of PC to evolve background knowledge toward contextualization and personalization. We demonstrated PC and its complementary nature to SC and CC by taking a concrete, real-world example of asthma management. The Internet of Things, often hailed as the next great phase of the Web, with its emphasis on sensing and actuation, will exploit all these three forms of computing.